%% file: neurips_2021.tex
\documentclass{article}

 \usepackage[preprint]{neurips_2021}

\usepackage[utf8]{inputenc} 
\usepackage[T1]{fontenc}    
\usepackage{hyperref}       
\usepackage{url}            
\usepackage{booktabs}       
\usepackage{amsfonts}       
\usepackage{nicefrac}       
\usepackage{microtype}      
\usepackage{xcolor}         
\usepackage{mathrsfs}

\usepackage{amsmath}
\usepackage{multirow}
\usepackage{makecell}
\usepackage{amsfonts,amssymb} 

\usepackage{natbib}
\usepackage[ruled]{algorithm2e}
\usepackage[]{algorithmic}
\usepackage{booktabs}
\usepackage{wrapfig}
\usepackage{enumitem}
\usepackage[position=b]{subcaption}

\usepackage{graphicx}
\usepackage{color}
\usepackage{pstricks}

\newrgbcolor{aasacolor}{.7 .1 .1}

\newrgbcolor{guancolor}{.6 .2 .0}

\newrgbcolor{francoiscolor}{.3 .2 .8}

\input{notations-2021.tex}
\title{Graph2Graph Learning\\ with Conditional Autoregressive Models}

\author	{%
	Guan Wang\\
	Dept. of Applied Mathematics and Computer Science\\
	Technical University of Denmark\\
	Lyngby, Denmark \\
	\texttt{guawa@dtu.dk} \\
	 \And
	 Francois Bernard Lauze \\
	 Dept. of Computer Science \\
	 University of Copenhagen \\
		Copenhagen, Denmark \\
	 \texttt{francois@di.ku.dk} \\
	 \And
	 Aasa Feragen \\
	 Dept. of Applied Mathematics and Computer Science \\
	 Technical University of Denmark \\
	 Lyngby, Denmark \\
	 \texttt{afhar@dtu.dk} \\
}

\begin{document}
	
\maketitle

\begin{abstract}
\input{abstract}
\end{abstract}

\section{Introduction}

\input{intro.tex}

\section{Method}
\label{sec:Method}

\input{method.tex}

\section{Experiments and Evaluation}
\label{sec:ExperimentsAndEvaluation}

\input{experiments.tex}
\section{Discussion and Conclusion}
\label{sec:DiscussionConclusion}

\input{discussion.tex}

\medskip

\small
\bibliographystyle{plainnat}
\bibliography{neurips_2021}

\end{document}

%% file: notations-2021.tex
\newcommand{\graph}{G}
\newcommand{\nodes}{V}
\newcommand{\edges}{E}
\newcommand{\adjmatrix}{A}
\newcommand{\real}{ \mathbb{R}}
\newcommand{\edgeAttrDim}{\gamma}

\newcommand{\node}{v}
\newcommand{\adjvec}{X}

\newcommand{\outAdjVec}{Y}
\newcommand{\outAdjVecs}{\textbf{Y}}
\newcommand{\outNodeNum}{m}

\newcommand{\srcGraphSeq}{\textbf{X}}
\newcommand{\trgGraphSeq}{\textbf{Y}}

\newcommand{\maxClique}{\graph^{*}}

\newcommand{\ie}{{\it{i.e., }}}

\newcommand{\context}{C}
\newcommand{\edgecontext}{c}
\newcommand{\nodeHiddenWeight}{\alpha}
\newcommand{\compatibilityFunc}{\phi}

\newcommand{\eNodeHiddenInv}{\overleftarrow{h}^{node}}				
\newcommand{\eNodeHiddenForw}{\overrightarrow{h}^{node}}				
\newcommand{\eNodeHidden}{h^{node}}									

\newcommand{\encEdgeRNN}{g_{edge}}
\newcommand{\encForwEdgeRNN}{\overrightarrow{g}_{edge}}

\newcommand{\eEdgeHiddenForw}{\overrightarrow{h}^{edge}}				

\newcommand{\eEdgeHidden}{h^{edge}}
\newcommand{\adjvecEmb}{\mathcal{X}}

\newcommand{\decEdgeRNN}{f_{edge}}
\newcommand{\decNodeRNN}{f_{node}}
\newcommand{\decDownEdgeRNN}{f_{edge,down}}

\newcommand{\decNodeHidden}{s^{node}}
\newcommand{\decEdgeHidden}{s^{edge}}
\newcommand{\decEdgeOutput}{o}
\newcommand{\outAdjVecEmb}{\mathcal{Y}}

\newcommand{\encForwNodeRNN}{\overrightarrow{g}_{node}}
\newcommand{\encInvNodeRNN}{\overleftarrow{g}_{node}}

\newcommand{\modelFunc}{\mathcal{M}}
\newcommand{\lossFuncAll}{\mathcal{L}}
\newcommand{\lossFuncSingle}{\ell}
\newcommand{\GTgraph}{\mathbb{Y}}

\newcommand{\setIJ}{\mathcal{I}}

%% file: abstract.tex
We present a graph neural network model for solving graph-to-graph learning problems. Most deep learning on graphs considers ``simple'' problems such as graph classification or regressing real-valued graph properties. For such tasks, the main requirement for intermediate representations of the data is to maintain the structure needed for output, i.e.~keeping classes separated or maintaining the order indicated by the regressor. However, a number of learning tasks, such as regressing graph-valued output, generative models, or graph autoencoders, aim to predict a graph-structured output. In order to successfully do this, the learned representations need to preserve far more structure. We present a conditional auto-regressive model for graph-to-graph learning and illustrate its representational capabilities via experiments on challenging subgraph predictions from graph algorithmics; as a graph autoencoder for reconstruction and visualization; and on pretraining representations that allow graph classification with limited labeled data.

%% file: intro.tex
Graphs are everywhere! While machine learning and deep learning on graphs have for long caught wide interest, most research continues to focus on relatively simple tasks, such as graph classification~\cite{ying2018hierarchical,xu2018powerful}, or regressing a single continuous value from graph-valued input~\cite{chen2019alchemy,yang2019analyzing,dwivedi2020benchmarkgnns,ok20similarity,bianchi21arma,zhang20dlgsurvey,wu21comprehensivesurvey}. While such tasks are relevant and challenging, they ask relatively little from the learned intermediate representations: For graph classification, performance relies on keeping classes separate, and for regressing a single real variable, performance relies on ordering graphs according to the output variable, but within those constraints, intermediate graph embeddings can shuffle the graphs considerably without affecting performance.

In this paper, we consider Graph2Graph learning, \ie problems whose input features and output predictions are both graphs. Such problems include both Graph2Graph regression and graph autoencoder models. For such problems, the model has to learn intermediate representations that carry rich structural information about the encoded graphs.

\paragraph{Related work.} While most existing work on graph neural networks (GNNs) centers around graph classification, some work moves in the direction of more complex output. Within chemoinformatics, several works utilize domain specific knowledge to predict graph-valued output, e.g.~ chemical reaction outcomes predicted via difference networks~\cite{jin2017predicting}. Similarly utilizing domain knowledge, junction trees are used to define graph variational autoencoders~\cite{junctionVAE} and Graph2Graph translation models~\cite{jin2018learning} designed specifically for predicting molecular structures, and interpretable substructures and structural motifs are used to design generative models of molecular Graph2Graphs in~\cite{jin2020multi} and~\cite{jin2020hierarchical}, respectively. Some work on encoder-decoder networks involving graph pooling and unpooling exist, but are only applied to node- and graph classification~\cite{gao2019graph,ying2018hierarchical}. More general generative models have also appeared, starting with the  Variational Graph Autoencoder~\cite{kipf2016variational} which is primarily designed for link prediction in a single large graph. In~\cite{li2020dirichlet}, clustering is leveraged as part of a variational graph autoencoder to ensure common embedding of clusters. This, however, leads to a strong dependence on the similarity metric used to define clusters, which may be unrelated to the task at hand.  In~\cite{you2018graphrnn}, a graph RNN, essentially structured as a residual decoder architecture, is developed by using the adjacency matrix to give the graph an artificial sequential structure. This generative model samples from a distribution fitted to a population of data, and does not generalize to graph-in, graph-out prediction problems. 

\textbf{We contribute} a model for Graph2Graph prediction with flexible modelling capabilities. Utilizing the graph representation from~\cite{you2018graphrnn}, we build a full encoder-decoder network analogous to previous work in sequence-to-sequence learning~\cite{sutskever2014sequence}. Drawing on tools from image segmentation, we obtain edge-wise probabilities for the underlying discrete graphs, as well as flexible loss functions for handling the class imbalance implied by sparse graphs. Our graph RNN creates rich graph representations for complex graph prediction tasks.  We illustrate its performance both for graph regression, as a graph autoencoder including visualization, and for unsupervised pretraining of graph representations to allow graph classification with limited labeled data.

%% file: method.tex
Our framework aims to generate new output graphs as a function of input graphs. Below, we achieve this by treating Graph2Graph regression as a sequence-to-sequence learning task by representing the graph as a sequence structure.

\subsection{Graph Representation}


Our framework takes as input a collection of undirected graphs with node- and edge attributes, denoted as $\graph = \{\nodes, \edges\}$, where $\nodes$ is the set of nodes and $\edges$ is the set of edges between nodes. Given a fixed node ordering $\pi$, a graph $\graph$ is uniquely represented by its attributed adjacency matrix 
$\adjmatrix \in \{\real^\edgeAttrDim\}^{n\times n}$, where $\edgeAttrDim$ is the attribute dimension for edges in $\graph$ and $n$ is the number of nodes. Moreover, $\adjmatrix_{i,j}$ is not $null$ iff $(\node_{i}, \node_{j}) \in \edges$.

\begin{figure}[t]
	\begin{minipage}[t]{0.6\linewidth}
		\centering
		\includegraphics[width=.9\linewidth]{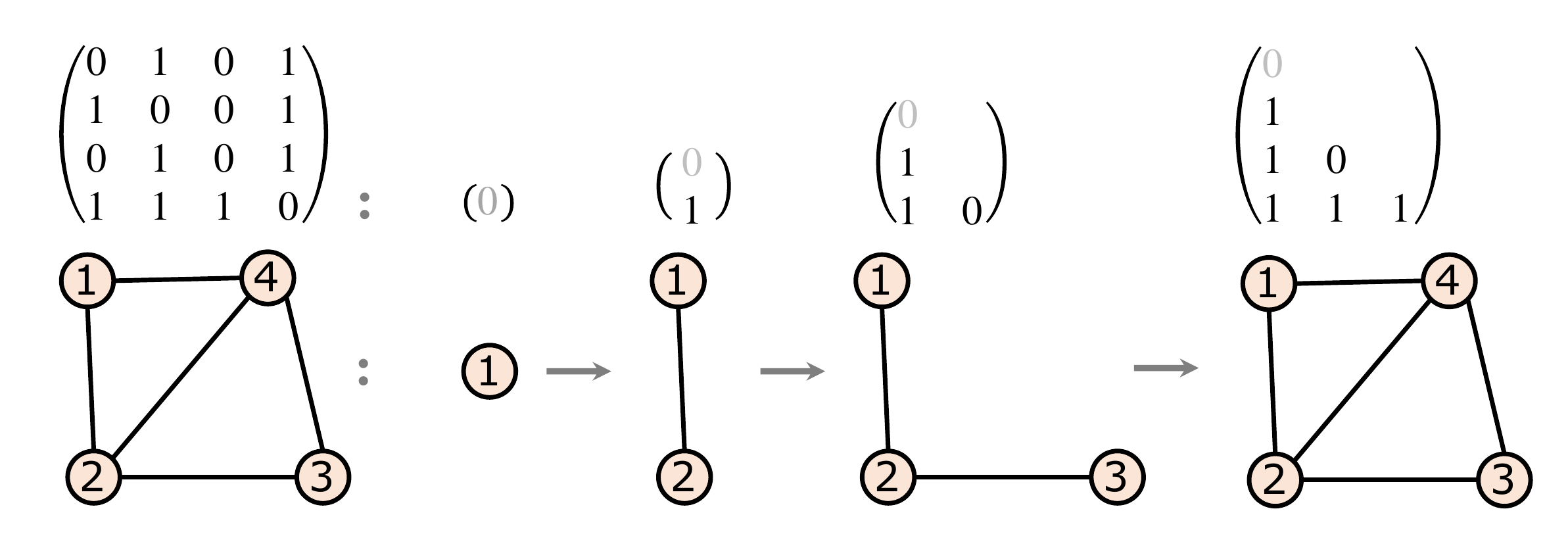} 
		\caption{Graph representation: a toy example.} 
		\label{fig:graph_representation}
	\end{minipage} 
	\begin{minipage}[t]{0.4\linewidth}
		\centering
		\includegraphics[width=.95\linewidth]{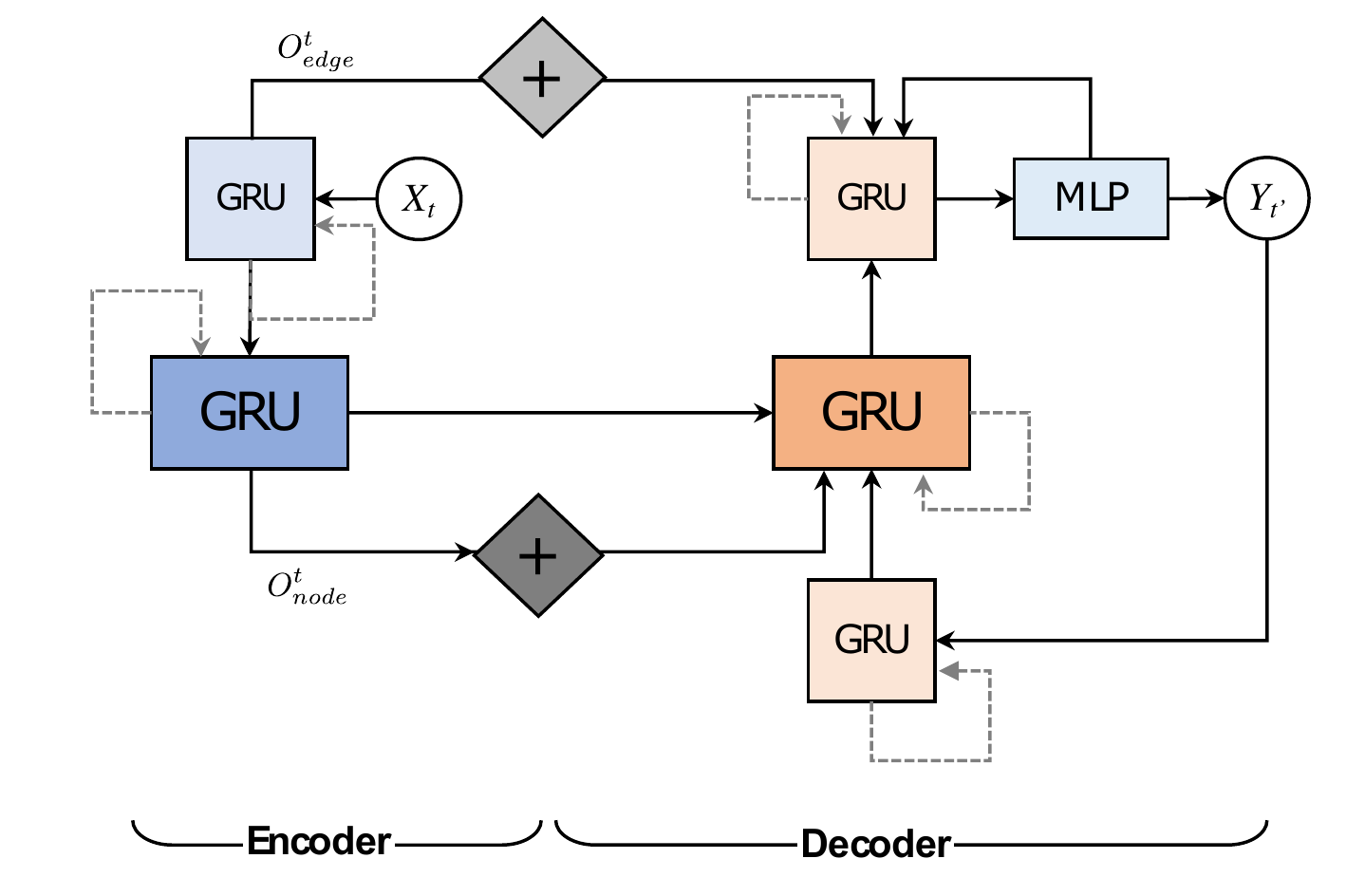} 
		\caption{Rolled representation of the proposed Graph2Graph network} 
		\label{fig:rolling_network_architecture}
	\end{minipage} 
	
\end{figure}

Consistent with the representation in~\cite{you2018graphrnn}~\cite{popova2019molecularrnn}, a graph $\graph$ will be represented as sequences of adjacency vectors $\{\adjvec_1, \adjvec_2,\cdots, \adjvec_n\}$ obtained by breaking $\adjmatrix$ up by rows. $\adjvec_i = (\adjmatrix_{i, i-1},\adjmatrix_{i, i-2},\cdots,\adjmatrix_{i,1})$ encodes the sequence of attributes of edges connecting the node $\node_i$ to its previous nodes $\{\node_{i-1}, \node_{i-2}, ..., \node_1\}$. The graph $\graph$ is thus transformed into a sequence of sequences, spanning across graph nodes $\nodes$ and edges $\edges$. A toy example (one graph with 4 nodes) is shown in Fig.~\ref{fig:graph_representation}. This representation allows us to build on graph representation learning for sequences.


\subsection{Graph2Graph Network Architecture}

\begin{figure}[t]
	\center
	\includegraphics[width=0.75\textwidth]{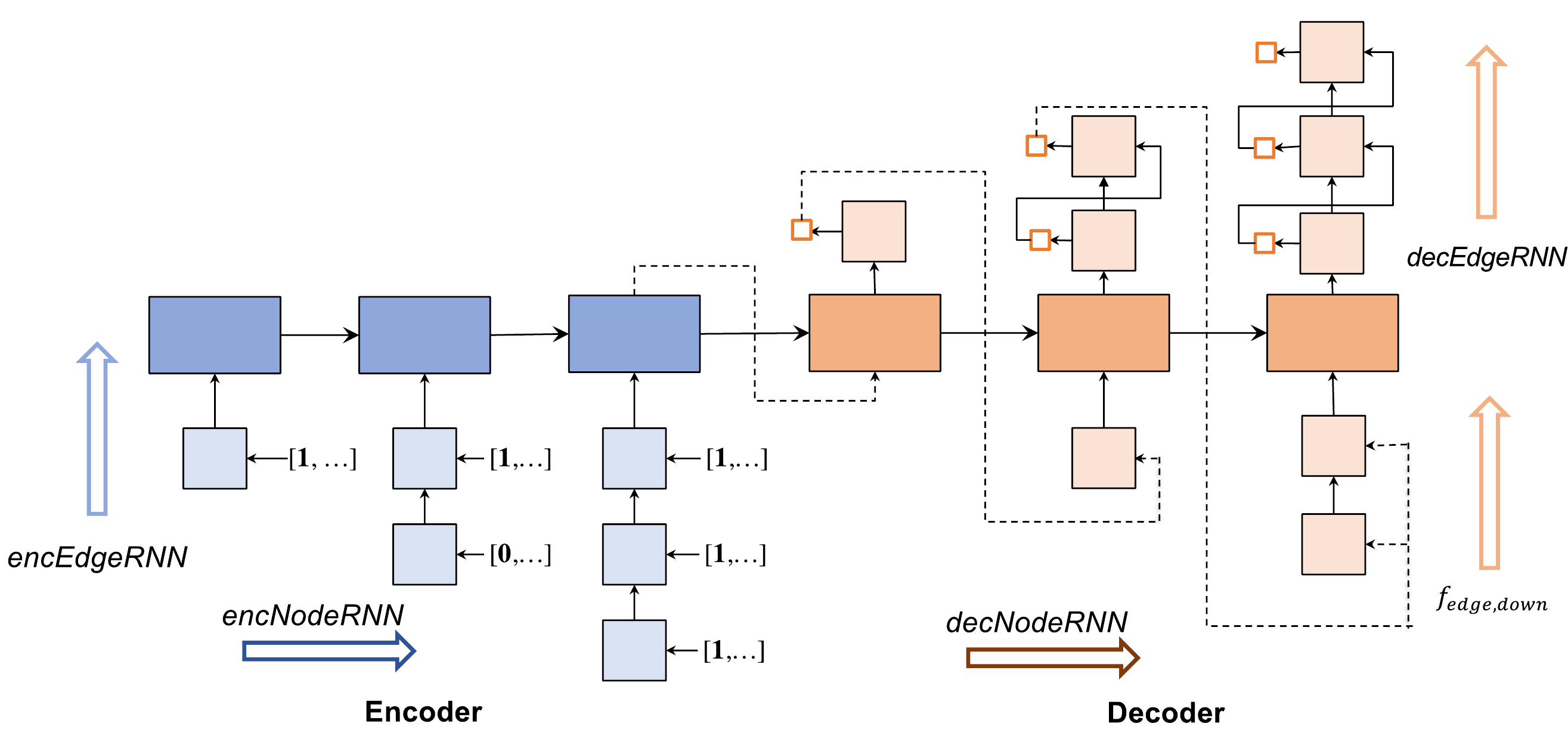} \hfil
	\gray{\vline}
	\includegraphics[width=0.2\textwidth]{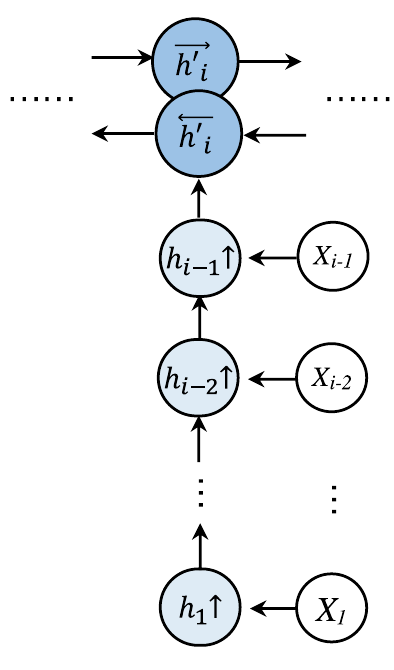}
	\caption{\textbf{Left:} Unrolled graphical representation of the proposed Graph2Graph network. \textbf{Right:} The framework that encodes an adjacency vector $\adjvec_i = \{\adjmatrix_{i,i-1}, \adjmatrix_{i,i-2},\cdots, \adjmatrix_{i,1}\}$ and fed into nodeRNN to help generate hidden state $\protect\eNodeHiddenForw_i$ and $\protect\eNodeHiddenInv_i$ at time $i$.}
	\label{fig:unrolling_network_architecture}
\end{figure}

Using the above graph representation, we propose a novel encoder-decoder architecture for Graph2Graph predictions building on previous models for sequence-to-sequence prediction~\cite{sutskever2014sequence}. At the encoding phase, graph information will be captured at edge level as well as node level, and analogously, the decoder will infer graph information at edge and node level in turn. A graphical illustration of proposed network is shown in Fig.~\ref{fig:rolling_network_architecture}, with its unrolled version in  Fig.~\ref{fig:unrolling_network_architecture}.


\subsubsection{Encoder}

The encoder aims to extract useful information from the input graph to feed into the decoding inference network. Utilizing the sequential representation of the input graph $G$, we use expressive recurrent neural networks (RNN) to encode $G$. The structure and attributes in $G$ are summarized across two levels of RNNs, each based on~\cite{sutskever2014sequence}: the node level RNN, denoted as \textit{encNodeRNN}, and the edge level RNN, denoted as \textit{encEdgeRNN}. For the \textit{encNodeRNN}, we apply bidirectional RNNs~\cite{schuster1997bidirectional} to encode information from both previous and following nodes, see Fig.~\ref{fig:unrolling_network_architecture} right subfigure. 


 The encoder network \textit{encEdgeRNN} reads elements in $X_i$ as they are ordered, \ie from $X_{i,1}$ to $X_{i,i-1}$, using the forward RNN $\encForwEdgeRNN$. Here, $\encForwEdgeRNN$ is a state-transition function, more precisely a Gated Recurrent Unit (GRU). The GRU $\encForwEdgeRNN$ produce a sequence of forward hidden states $(\eEdgeHiddenForw_{i,1},\eEdgeHiddenForw_{i,2}, \cdots, \eEdgeHiddenForw_{i,i-1})$. Then we pick $\eEdgeHiddenForw_{i,i-1}$ to be the context vector of $X_i$, and input it into \textit{encNodeRNN}
 

The encoder network \textit{encNodeRNN} is a bidirectional RNN, which receives input from  \textit{encEdgeRNN} and transmits hidden states over time, as shown in Figs.~\ref{fig:rolling_network_architecture} and~\ref{fig:unrolling_network_architecture}. This  results in concatenated hidden states $\eNodeHidden_{i} = \eNodeHiddenForw_i\|\eNodeHiddenInv_{n-i}$. The final hidden state $\eNodeHiddenForw_{n}$ and $\eNodeHiddenInv_{1}$ are concatenated and used as the initial hidden state for the decoder.

\subsubsection{Decoder}
Once the input graph is encoded, the decoder seeks to infer a corresponding output graph for a given problem. To help understand how the decoder works, we consider a particular graph prediction problem motivated by an NP hard problem from graph algorithms, namely the identification of a \emph{maximum clique}. In this application, we take a graph as input and aim to predict which nodes and edges belong to its maximum clique. Thus, the decoder predicts whether to retain or discard the next edge given all previous nodes, predicted edges and the context vector of the input graph. 

The decoder defines a probability distribution $p(\outAdjVecs)$, where $\outAdjVecs = (\outAdjVec_0,\outAdjVec_1,\dots,\outAdjVec_{m-1}) $ is the sequence of adjacency vectors that forms the output as predicted by the decoder. Here, $p(\outAdjVecs)$ is a joint probability distribution on the adjacency vector predictions and can be decomposed as the product of ordered conditional distributions:
\begin{equation}
	\setlength{\abovedisplayskip}{3pt}
	\setlength{\belowdisplayskip}{3pt}
	p(\outAdjVecs) = \prod_{i=0}^{\outNodeNum-1}p(\outAdjVec_i|\outAdjVec_0,\outAdjVec_1,\dots,\outAdjVec_{i-1}, \context), 
	\label{eq:dec_nodeProbDistribution}
\end{equation}
\noindent where $\outNodeNum$ is the node number of the predicted graph,  and $\context$ is the context vector used to facilitate predicting $\outAdjVec_i$, as explained in Sec.~\ref{sec:attention} below. Denote $p(\outAdjVec_i|\outAdjVec_0,\outAdjVec_1,\dots,\outAdjVec_{i-1}, \context)$ as $p(\outAdjVec_i|\outAdjVec_{<i}, \context)$ for simplication.

In order to maintain edge dependencies in the prediction phase, $p(\outAdjVec_i|\outAdjVec_{<i}, \context)$ is factorized as composed as a product of conditional probabilities. 
\begin{equation}
	\setlength{\abovedisplayskip}{3pt}
	\setlength{\belowdisplayskip}{3pt}
	p(\outAdjVec_i|\outAdjVec_{<i}, \context) = \prod_{k=0}^{i-1}p(\outAdjVec_{i,k}|\outAdjVec_{i,<k},\edgecontext; \outAdjVec_{<i}, \context)
	\label{eq:dec_edgeProbDistribution}
\end{equation}
\noindent where $\edgecontext$ is the edge context vector for predicting $\outAdjVec_{i,k}$.

The cascaded relation between Equation~\eqref{eq:dec_nodeProbDistribution} and Equation~\eqref{eq:dec_edgeProbDistribution}, is reflected in practice by their approximation by the two cascaded RNNs called \textit{decNodeRNN} and \textit{decEdgeRNN}. Here, \textit{decNodeRNN} transits graph information from node $i-1$ to node $i$ hence generating a node (see Eq.~\eqref{eq:decNodeRNN}), and \textit{decEdgeRNN} generates edge predictions for the generated node (Eq.~\eqref{eq:decEdgeRNN}). The output of each \textit{decNodeRNN} cell will serve as an initital hidden state for \textit{decEdgeRNN}. Moreover, at each step, the output of the \textit{decEdgeRNN} will be fed into an MLP head with sigmoid activation function, generating a probability of keeping this edge in the output graph (Eq.~\eqref{eq:decEdgeMLP}). This use of sigmoid activation is similar to its use in image segmentation, where a pixel-wise foreground probability is obtained from it.
\begin{align}
	\setlength{\abovedisplayskip}{3pt}
	\setlength{\belowdisplayskip}{3pt}
	\label{eq:decNodeRNN}
	&\decNodeHidden_i = \decNodeRNN(\decNodeHidden_{i-1}, \decDownEdgeRNN(\outAdjVec_{i-1}), \context_i )		\\	
	\label{eq:decEdgeRNN}
	&\decEdgeHidden_{i,j}=\decEdgeRNN(\decEdgeHidden_{i,j-1}, emb(\outAdjVec_{i,j-1}), \edgecontext_{i,j}) 		\\
	\label{eq:decEdgeMLP}	
	&\decEdgeOutput_{i,j} = \text{MLP}(\decEdgeHidden_{i,j})
\end{align}
Our decoder has some features in common with the graph RNN~\cite{you2018graphrnn}. We extend this by utilizing attention based on the decoder, as well as by adding in extra GRUs in the $\decDownEdgeRNN$, encoding the predicted adjacency vector from the previous step in order to improve model expressive capability.

\begin{algorithm}[t]
	
	\caption{\label{alg:overall algorithm}Graph to Graph learning algorithm.}
	
	\textit{Input:} graph $\srcGraphSeq = (\adjvec_1,\dots, \adjvec_n)$.
	
	\textit{Output:} graph $\trgGraphSeq = (\outAdjVec_1,\dots, \outAdjVec_m)$. 
	
	\begin{algorithmic}[1]
		
		\STATE For each $\adjvec_i, i=1,\dots,n$, use edge RNN $\encEdgeRNN$ to calculate their encodings$\{\adjvecEmb_1,\dots,\adjvecEmb_n\}$ and hidden states $\eEdgeHidden_{i, j}$.
		\STATE Forward node RNN $\encForwNodeRNN$ reads $\{\adjvecEmb_1,\dots,\adjvecEmb_n\}$ and calculates forward node hidden states $\{\eNodeHiddenForw_1,\dots,\eNodeHiddenForw_n\}$
		\STATE Reverse node RNN $\encInvNodeRNN$ reads $\{\adjvecEmb_n,\dots,\adjvecEmb_1\}$ and calculates reverse node hidden states $\{\eNodeHiddenInv_1,\dots,\eNodeHiddenInv_n\}$ 
		
		\STATE Final hidden state $\eNodeHidden_i = \textsc{Concat}(\eNodeHiddenForw_i, \eNodeHiddenInv_{n-i+1})$
		
		\STATE Set initial hidden state of decoder node RNN as $\decNodeHidden_1 = \eNodeHidden_n$, initial edge sequence encoding $\outAdjVecEmb_0 = \mathrm{SOS}, i=1$ 
		
		\STATE While $i<=m$ do
		\begin{itemize}
			\item $\decDownEdgeRNN$ encodes $\outAdjVec_{i-1}$ to be $\outAdjVecEmb_{i-1}$
			\item $\decNodeHidden_i = \decNodeRNN(\decNodeHidden_{i-1},\outAdjVecEmb_{i-1}, \textup{NodeAttn}(\eNodeHidden_{1:n},\decNodeHidden_{i-1} ))$
			
			\item $\decEdgeHidden_{i,0} = \decNodeHidden_i, j=1, \outAdjVec_{i,0} = \mathrm{SOS_{decEdge1}}$
			\item While $j<=i$ do
			\begin{itemize}
				\item $\decEdgeHidden_{i,j} = \decEdgeRNN(\decEdgeHidden_{i,j-1}, \outAdjVec_{i,j-1}, \text{EdgeAttn}(\decEdgeHidden_{i,k}, \eEdgeHidden_{:,:}))$
				\item $\outAdjVec_{i,j} = \text{MLP}(\decEdgeHidden_{i,j})$
				\item $j\leftarrow j+1$ 
			\end{itemize}
			\item $i\leftarrow i+1$
		\end{itemize}
		\STATE Return predicted graph sequence $\outAdjVec_{1:m}$
	\end{algorithmic}
	
	\label{alg:overall overall algorithm}
\end{algorithm}

\subsubsection{Attention mechanism} \label{sec:attention}

Seen from Eq.~\eqref{eq:dec_nodeProbDistribution}~\eqref{eq:dec_edgeProbDistribution}, the final edge probabilities are conditioned on a node context vector $\context$, as well as an edge context vector $\edgecontext$. We derive these following~\cite{bahdanau2014neural}:

The node context vector $\context_i$, is defined from the hidden vectors $(\eNodeHidden_1, \eNodeHidden_2, \dots, \eNodeHidden_n)$ generated by the encoder. Each hidden vector $\eNodeHidden_i$ has captured global information from the graph $\graph$, with a main focus on the $i$-th node. Now, the $i^{th}$ node context vector $\context_i$ is computed as the weighted sum across all these hidden vectors:
\begin{equation}
	\setlength{\abovedisplayskip}{3pt}
	\setlength{\belowdisplayskip}{3pt}
	\context_i = \sum_{j=1}^{n}\nodeHiddenWeight_{i,j}\eNodeHidden_j.
	\label{eq:nodeAttnComputation}	
\end{equation}
The weight assigned to $\eNodeHidden_j$ is computed as the normalized compatibility of $\eNodeHidden_j$ with $\decNodeHidden_{i-1}$:
\begin{equation}
	\setlength{\abovedisplayskip}{3pt}
	\setlength{\belowdisplayskip}{3pt}
	\nodeHiddenWeight_{i,j} = \frac{\exp(\compatibilityFunc(\decNodeHidden_{i-1}, \eNodeHidden_j))}
	{\sum_{k=1}^{n}\exp(\compatibilityFunc(\decNodeHidden_{i-1}, \eNodeHidden_k))}
	\label{eq:nodeAttnWeightComputation}							
\end{equation}
\noindent where $\compatibilityFunc$ is a compatibility function, parameterized as a feedforward neural network and trained jointly along with other modules in the whole model.

The computation of the edge level context $\edgecontext_{i,j}$ follows the same scheme. The overall framwork is summarized in Algorithm~\ref{alg:overall algorithm}.

%% file: experiments.tex
Next, we show how our model can be applied in a wide range of tasks, including Graph2Graph regression applied as a heuristic to solve challenging graph algorithmic problems; representation learning as a graph autoencoder; and utilizing the graph autoencoder to learn semantic representations for graph classification with limited labeled data.

\paragraph{General experimental setup.} All models were trained on a single NVIDIA Titan GPU with 12GB memory, using the Adam~\cite{Kingma2015}  optimizer with learning rate $\in \{0.01, 0.003\}$ and batch size $\in \{64,128\}$.

\paragraph{Loss Function.} For all Graph2Graph learning tasks, we used the Focal loss~\cite{lin2017focal} function known from segmenting images with unbalanced classes, analogous to the relatively sparse graphs. Note that image segmentations and binary graphs are similar as target objects, in the sense that both consist of multiple binary classifications.

For an input graph $\srcGraphSeq$, a ground truth output graph $\GTgraph$ and a Graph2Graph network $\modelFunc$, the loss $\lossFuncSingle_{i,j}$ for the edge between node $i$ and its $j$-th previous node is :
\begin{equation}
	\setlength{\abovedisplayskip}{3pt}
	\setlength{\belowdisplayskip}{3pt}
	\lossFuncSingle_{i,j} = -(1-p^t_{i,j})^\gamma \log(p^t_{i,j}), 
	\label{eq:lossFuncSingle}
\end{equation}

\noindent where $p^t_{i,j}$ denotes likelihood of \emph{correct prediction}, and $\gamma>0$ is a hyperparameter used to reduce relative loss for well classified edges. In our experiment $\gamma=2$.

The final loss is a sum of edge-wise losses: $\lossFuncAll(\modelFunc(\srcGraphSeq), \GTgraph) = \sum_{(i, j)\in\setIJ}\lossFuncSingle_{i,j}$, where $\setIJ$ is the set of node index pairs. This varies depending on application: For the maximum clique prediction (Sec.~\ref{subsec:maxCliquePrediction}), we have $\forall a \in \setIJ, \srcGraphSeq_a = 1$, restricting to predicting subgraphs of the input. For the autoencoder (Sec.~\ref{sec:ae} and~\ref{sec:graphClassification}), on the other hand, $\setIJ$ contains all index pairs in \srcGraphSeq.

\begin{table}[t]
	\center
	{
		\caption{\label{tab:datasets_3tasks}Summary of datasets used in our experiments on three applications.}
		\resizebox{\textwidth}{!}
		{ 
			\setlength{\tabcolsep}{4pt}{
				\begin{tabular}{@{}llcccccc@{}}
					\toprule
					  ~  &\textbf{Dataset} 	&\# Graphs		&\# Node (avg)		&\# Node (max) &\# Training &\# Validation &\# Test \\
					\midrule
					\multirowcell{3}{Sec.~\ref{subsec:maxCliquePrediction}\\ \textbf{Maximum Clique}}	&DBLP\_v1 &{$14488$} &{$11.7$} &{$39$} &{$8692$}  &{$2898$}	&{$2898$}	\\
						~ &IMDB-MULTI		    &{$1500$} 	 &{$13.0$}	  &{$89$}	&{$900$}	  &{$300$}  &{$300$}	\\
						~ &deezer\_ego\_nets    &{$2558$}    &{$16.7$}	  &{$30$}   &{$1552$}	  &{$518$}	 &{$518$}	\\
					\midrule
					\multirowcell{3}{Sec.~\ref{sec:ae}\\ \textbf{AutoEncoder}}	&DBLP\_v1 	&{$19455$}	&{$10.5$}  &{$39$}	&{$11673 $} 	&{$3891 $}	&{$3891$}	\\
						~ &IMDB-MULTI		&{$1500$} 	&{$13.0$}	&{$89$}	&{$900$}	&{$300$}  	&{$300$}	\\
						 ~ 	&MUTAG    		&{$188$}    &{$17.9$}	  &{$28$}   &{$112$}	  &{$38$}	 &{$38$}	\\
	
					\midrule
					\multirowcell{3}{Sec.~\ref{sec:graphClassification}\\ \textbf{Classification}} &DBLP\_v1 	&{$19455$}	&{$10.5$}  &{$39$}	&{$11673 $} &{$3891 $}	&{$3891$}	\\
					 ~  &IMDB-MULTI    	&{$1500$}    &{$13.0$}	  &{$89$}   	&{$1200$}	  &{$150$}	 &{$150$}  \\
					 
					 ~ 	&AIDS    		&{$2000$}    &{$15.7$}	  &{$\emph{50}$}   &{$1600$}	  &{$200$}	 &{$200$}	\\
					~ 	&NCI1		    &{$4100$} 	 &{$29.9$}	  &{$\emph{50}$}	&{$3288$}	  &{$411$}  &{$411$}	\\
					~ 	&IMDB-BINARY   	&{$1000$}    &{$19.8$}	  &{$\emph{100}$}   &{$800$}	  &{$100$}	 &{$100$} \\
					\bottomrule
					
				\end{tabular}
			}
		}
	}
\end{table}

\begin{table}[t]
	\center
	\setlength\tabcolsep{10pt}
	{
		\caption{\label{tab:maxClique_acc_iou}Results of maximal clique prediction in term of accuracy(\%) and edge IOUs(\%) on DBLP\_v1, IMDB-MULTI and deezer\_ego\_nets dataset. OOM means out of memory even if batch size is 1, Graph2Graph denotes our proposed model. Results are based on a random split.}
		\resizebox{\textwidth}{!}
		{
			\begin{tabular}{@{}lcccccc@{}}
				\toprule
				~ & \multicolumn{2}{c}{DBLP\_v1}& \multicolumn{2}{c}{IMDB-MULTI}& \multicolumn{2}{c}{deezer\_ego\_nets} \\
				\midrule
				
				\textbf{Models} 	& Accuracy	&edge IoU	& Accuracy	&edge IoU		& Accuracy 	& edge IoU 			 \\	
				\cmidrule(r){1-1}		\cmidrule(r){2-3}  	\cmidrule(r){4-5}  			\cmidrule(r){6-7}
				MLP 				&$85.0$  & $93.8$ 	& $ 61.0 $ & $85.7 $ & $33.2$ & $66.7$ \\
				
				GRU w/o Attn 		&$85.96$ & $95.47$ 		&$54.33$  &$79.82$  &$42.86$ & $69.76 $\\
				
				GRU with Attn 		&\multicolumn{2}{c}{>3days} &\multicolumn{2}{c}{OOM}& $46.53$& $76.75$ \\
				
				
				Graph2Graph 		&$\textbf{95.51}$ & $\textbf{97.43}$  	&  $\textbf{82.3}$ & $\textbf{92.5}$ & $\textbf{58.5}$ & $\textbf{81.8}$ \\
				
				\bottomrule
			\end{tabular}
		}
	}
\end{table}

\begin{table}[!t]
	\center
	{
		\caption{\label{tab:ablation_maxClique_acc_iou}Ablation study on maximal clique prediction in term of accuracies and edge IOUs on DBLP\_v1, IMDB-MULTI and deezer\_ego\_nets dataset.  Graph2Graph denotes our proposed model. Results are computed on a random split}
		\resizebox{\textwidth}{!}
		{
			\begin{tabular}{@{}lcccccc@{}}
				\toprule
				~	& \multicolumn{2}{c}{DBLP\_v1}& \multicolumn{2}{c}{IMDB-MULTI}& \multicolumn{2}{c}{deezer\_ego\_nets} \\
				\cmidrule(r){2-3}  				\cmidrule(r){4-5}  			\cmidrule(r){6-7}
				\textbf{Models}				& Accuracy	&edge IoU		& Accuracy	&edge IoU& Accuracy & edge IoU 			 \\	
				\midrule
				Graph2Graph w/o NodeAttn	 &$93.48$ & $96.15$   &$63.67$ &$83.21$  &$50.39$ &$78.90$  \\
				
				Graph2Graph w/o EdgeAttn	 &$94.58$ & $96.58$   &$80.00$ &$90.51$  &$56.76$ &$\textbf{82.90}$\\
				
				Graph2Graph full			 &$\textbf{95.51}$  &$\textbf{97.43}$ 	&  $\textbf{82.3}$ & $\textbf{92.5}$  & $\textbf{58.5}$ & $81.8$   \\
				
				\bottomrule
			\end{tabular}
		}
	}
\end{table}

\begin{figure}[t]
	\begin{minipage}[b]{0.45\linewidth}
		\centering
		\includegraphics[width=.9\linewidth]{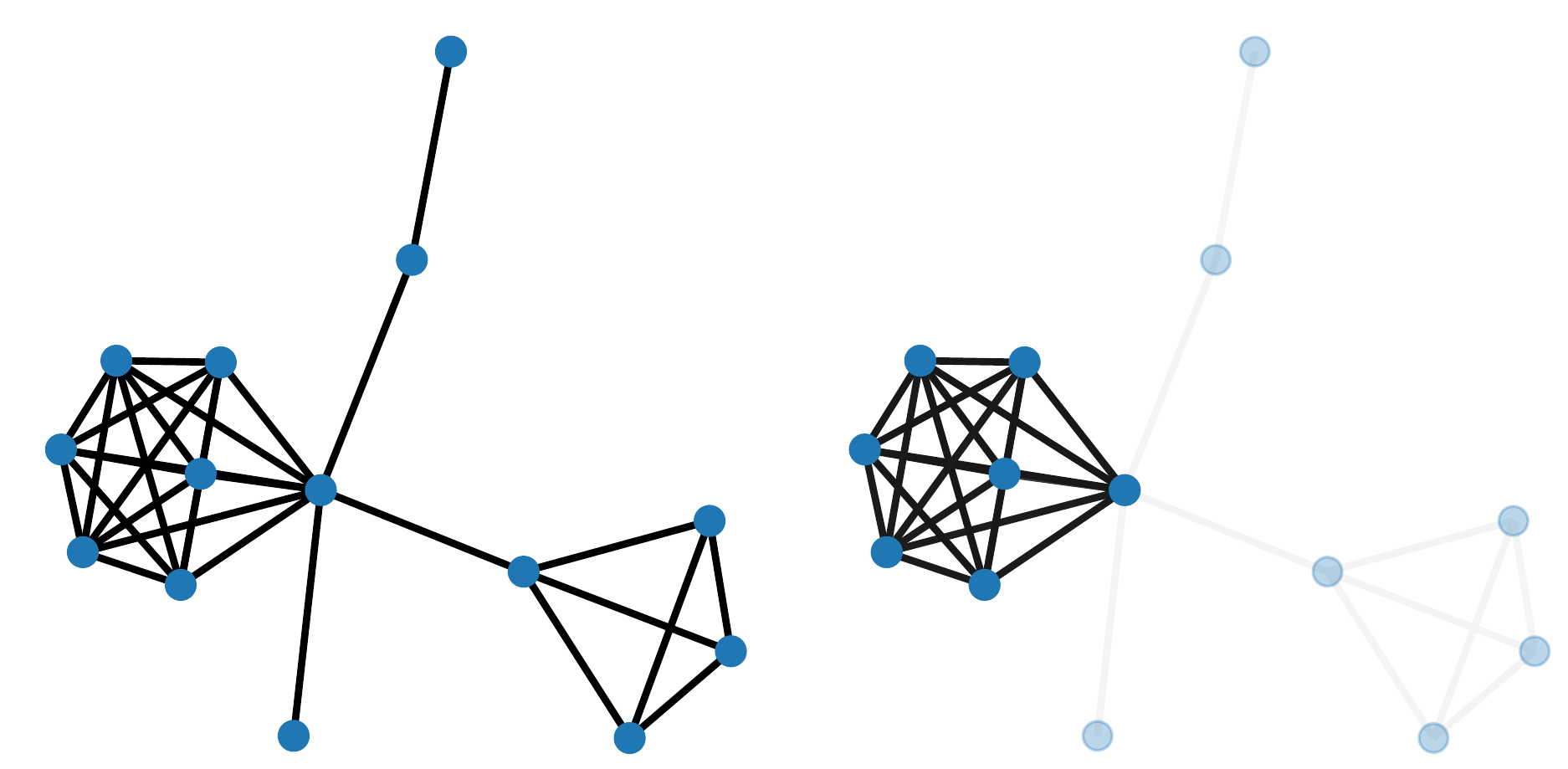}  
		\label{fig:maxClique_example_1}
		\subcaption{example from DBLP\_v1}
	\end{minipage}
	\gray{\vline}
	\begin{minipage}[b]{0.45\linewidth}
		\centering
		\includegraphics[width=.9\linewidth]{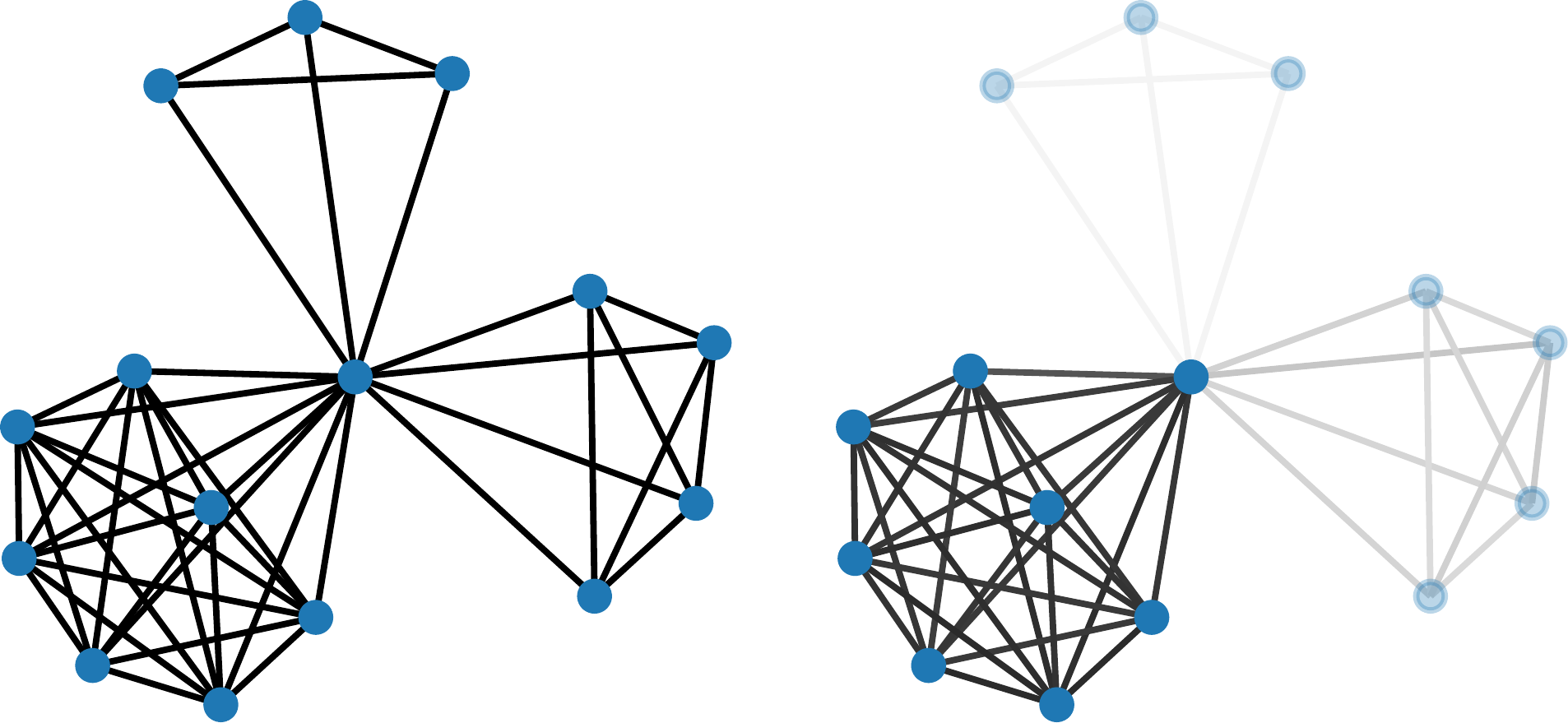}
		\subcaption{example from IDMB\_MULTI} 
		\label{fig:maxClique_example_2}
	\end{minipage}%
	\begin{minipage}[b]{0.04\linewidth}
		\centering
		\includegraphics[width=.9\linewidth]{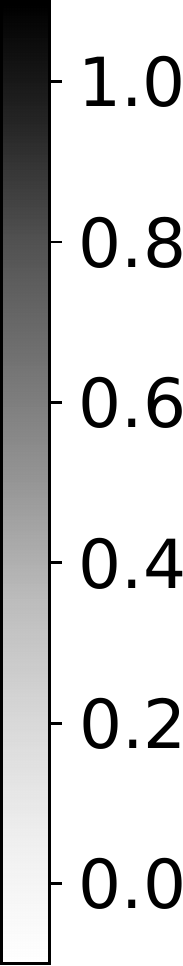}
		\label{fig:maxClique_example_colorBar}
	\end{minipage}
	\caption{Examples of maximal clique predictions; edge probability shown in colorbar.}
	\label{fig:maxClique_visualization}
\end{figure}

\subsection{Solving graph algorithmic problems via Graph2Graph regression}\label{subsec:maxCliquePrediction}

Many important graph problems are NP complete~\cite{bomze1999maximum, alekseev2007np}, enhancing the need for efficient heuristics. While most existing heuristics are deterministic in their nature, and hence also in the mistakes that they make, a neural network would be trained to perform the task at hand for a particular data distribution. This gives possibilities for improving both quality, tailoring the predictions to the data, and computational complexity, which is low at test time.

Here, we train the Graph2Graph architecture to predict maximum cliques, an NP complete problem, and illustrate its performance on several benchmark datasets.

\paragraph{Problem Definition and settings} The Maximal Clique (MC) is the \textit{complete} subgraph $\maxClique$ of a given graph $\graph$, which contains the maximal number of nodes. 
To reduce computational time, we employed a fixed attention mechanism for both node level and edge level by setting $\nodeHiddenWeight_{i,j} =1, \text{if } i=j$.

\paragraph{Data.}
Our experiments are carried out using the datasets DBLP\_v1~\cite{pan2013graph}, IMDB-MULTI~\cite{yanardag2015deep}, and deezer\_ego\_nets~\cite{rozemberczki2020api} from the TUD graph benchmark database~\cite{tud}. Graphs whose maximum clique contained less than 3 nodes were excluded, and for the deezer\_ego\_nets, we excluded those graphs that had more than 30 nodes, giving the dataset statistics shown in Table~\ref{tab:datasets_3tasks}. In each graph dataset, 60\% were used for training, 20\% for validation and 20\% for test.

\paragraph{Results.}
Performance was quantified both in terms of accuracy and Intersection over Union (IoU) on correctly predicted edges. While the former counts completely correctly predicted subgraphs, the latter quantifies near-successes, analogous to its use in image segmentation. We compare our performance with other similar autoencoder architectures with backbones as MLP, GRU with~\cite{bahdanau2014neural} and without Attention~\cite{cho2014learning}, by flattening a graph as one-level sequence. The results found in Table~\ref{tab:maxClique_acc_iou} clearly show that the Graph2Graph architecture outperforms the alternative models. This high performance is also illustrated in Fig.~\ref{fig:maxClique_visualization}, which shows visual examples of predicted maximal cliques. These are illustrated prior to thresholding, with edge probabilities indicated by edge color. More examples are found in the Supplementary Material.



\paragraph{Ablation Study.} By removing node level attention (\textit{Graph2Graph w/o NodeAttn}) and  edge level attention (\textit{Graph2Graph w/o EdgeAttn}) from the original model, we investigate the components' contributions to the MC prediction; see Table~\ref{tab:ablation_maxClique_acc_iou} for results. We see that Graph2Graph Models have better performance over Graph2Graph without NodeAttn and that without EdgeAttn on all three datasets under both metrics, except the edgeIoU on deezer\_ego\_nets. These results demonstrate the contributions of attention mechanism at node level and edge level to performance improvement.

\begin{figure}[!t]
	\centering
	\includegraphics[width=\linewidth]{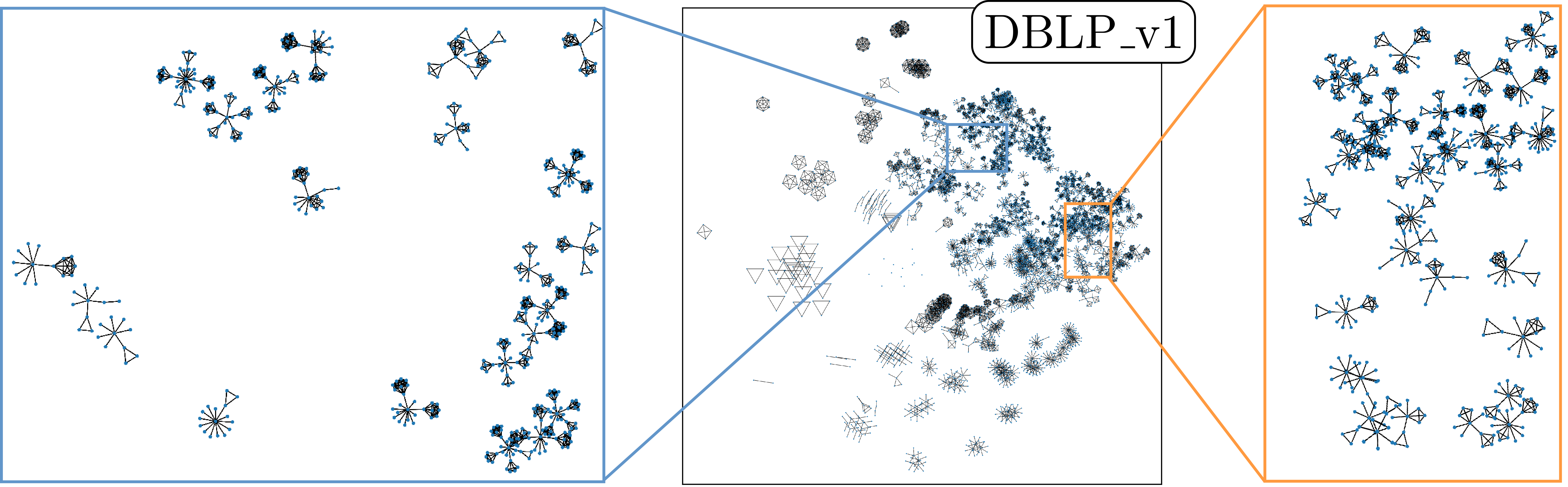}
	\caption{Latent representation for DBLP\_v1 test set.}
	\label{fig:dblp_tsne}
\end{figure}

\begin{figure}[!t]
	\centering
	\includegraphics[width=\linewidth]{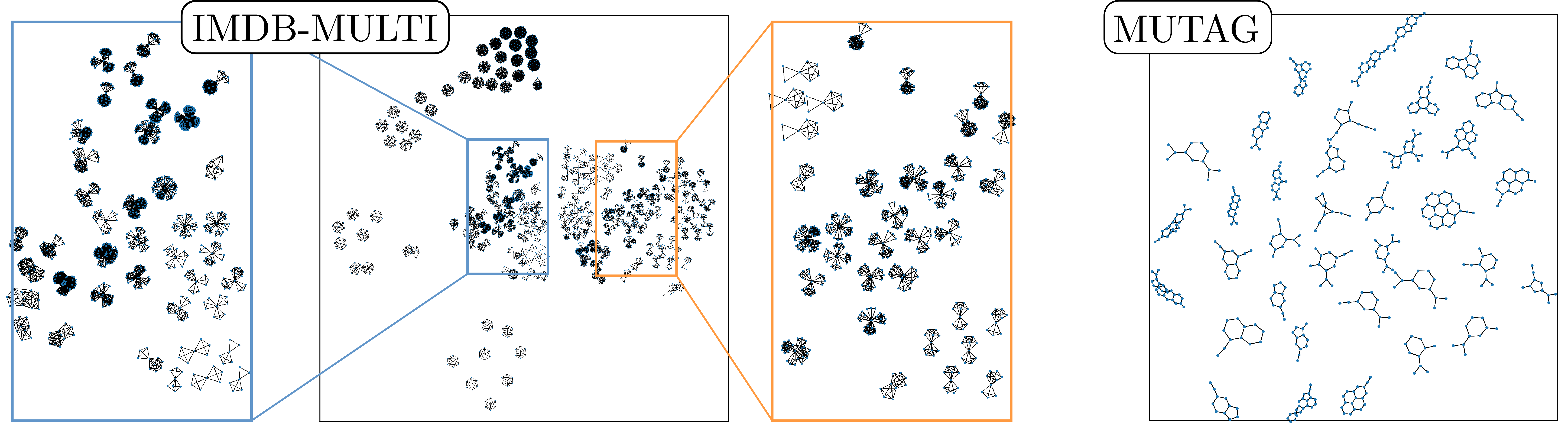}
	\caption{Latent representation for test sets of IMDB-MULTI (left) and MUTAG (right).}
	\label{fig:imdb_mutag_tsne}
\end{figure}

\subsection{Graph autoencoder via Graph2Graph prediction} \label{sec:ae}

It is well known from image- and language processing~\cite{cho2014learning}~\cite{sutskever2014sequence} that encoder-decoder networks often learn semantically meaningful embeddings when trained to reproduce their input as an autoencoder. We utilize the encoder-decoder structure of our proposed Graph2Graph model to train the network as a graph autoencoder, mapping graphs to continuous context vectors, and back to an approximation of the original graph.



\paragraph{Problem definition and settings.} Given input graphs $G$, train the Graph2Graph network $\modelFunc \colon \graph \mapsto H$ for the prediction $H = \hat{\graph}$ to reconstruct $\graph$ as well as possible. The encoder uses single directional RNNs for both edgeRNN and nodeRNN, \ie $\eEdgeHidden_i = \eEdgeHiddenForw_i$, $\eNodeHidden_i = \eNodeHiddenForw_i$. We use no edge context $\edgecontext$, and constrain all node contexts $\context$ as $\eNodeHidden_n$ to obtain more compressed encodings. The resulting $\eNodeHidden_n$ serves as a latent representation of $\graph$ in a learned latent space, which in our experiments has dimension 128.

\paragraph{Data.} We use the full TU~\cite{tud} datasets DBLP\_v1, IMDB-MULTI and MUTAG~\cite{debnath1991structure}, using 60/20/20\% for training/validation/test, respectively; see Table~\ref{tab:datasets_3tasks} for details.

\begin{figure}[t]
	\begin{minipage}[b]{0.4\linewidth}
		\centering
		\includegraphics[width=.99\linewidth]{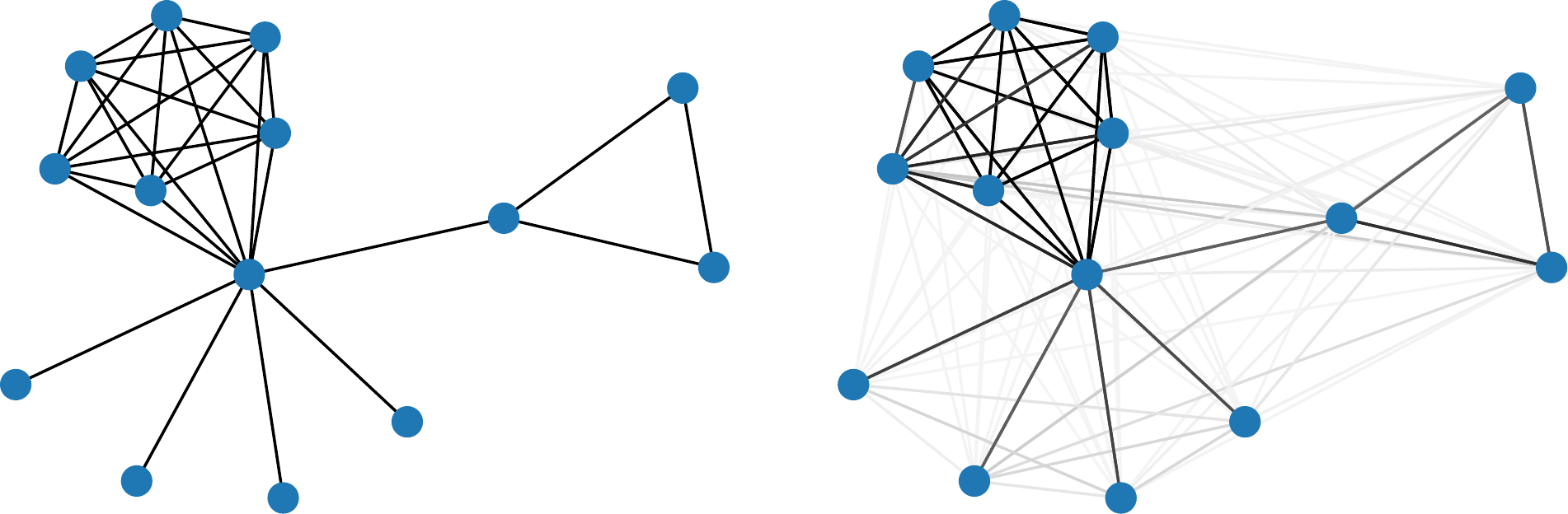}  
		\label{fig:graphReconstruction_1}
		\subcaption{sample from DBLP\_v1}
	\end{minipage}
	\gray{\vline}
	\begin{minipage}[b]{0.4\linewidth}
		\centering
		\includegraphics[width=.99\linewidth]{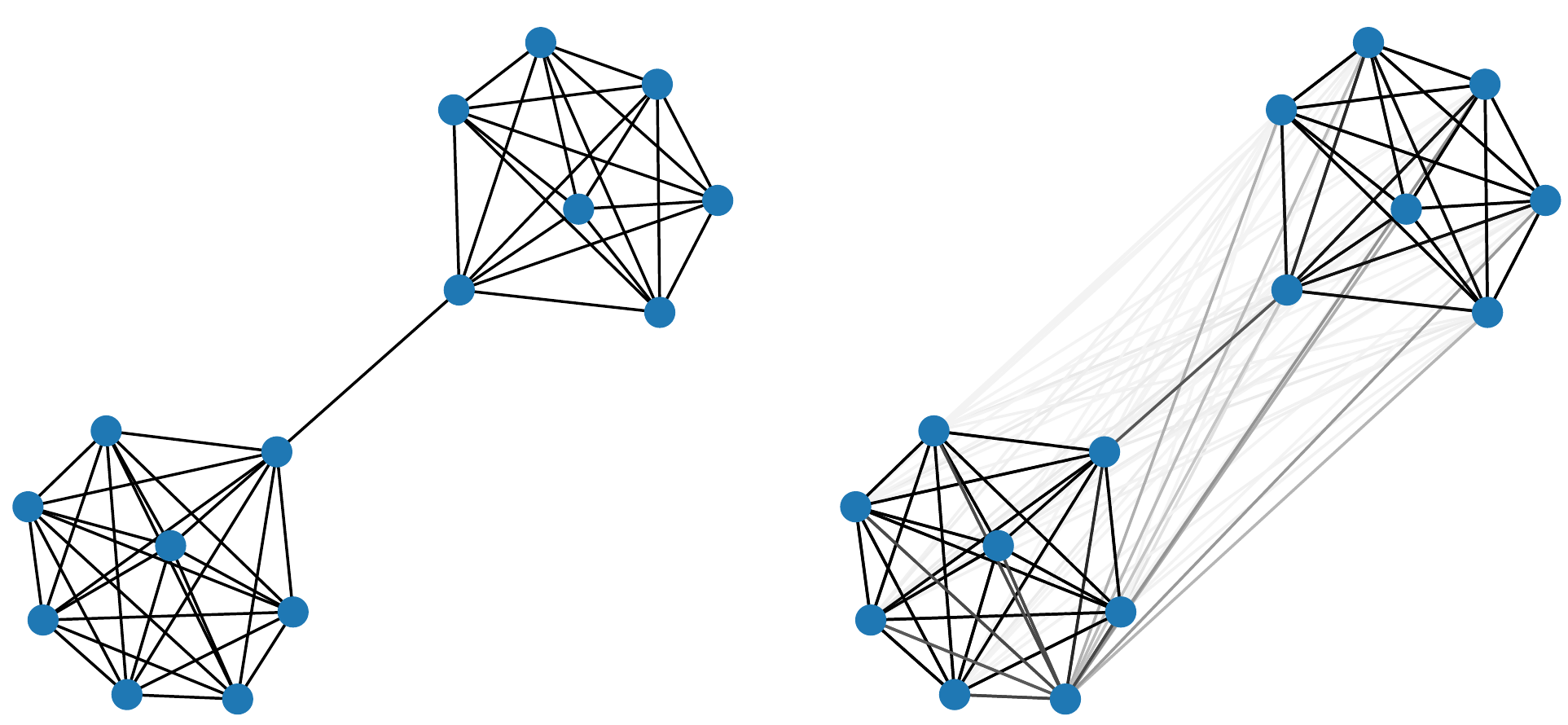}  
		\label{fig:graphReconstruction_2}
		\subcaption{sample from DBLP\_v1}
	\end{minipage}%
	\hfil
	\begin{minipage}[b]{0.04\linewidth}
		\centering
		\includegraphics[width=.9\linewidth]{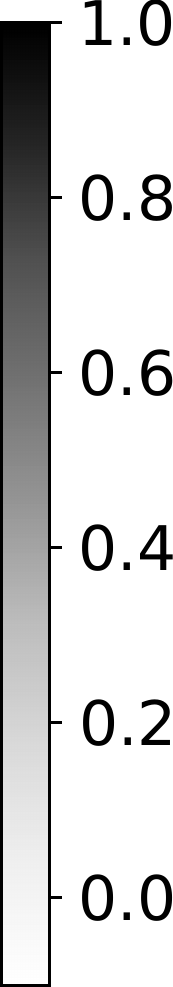}
		\label{fig:graphReconstruction_colorBar}
	\end{minipage}
	\caption{Autoencoder reconstructions from DBLP\_v1. Edge probability in grayscale (see colorbar).}
	\label{fig:graphReconstruction}
\end{figure}

\paragraph{Results.} A visual comparison of original and reconstructed graphs is found in Fig.~\ref{fig:graphReconstruction}, and Figs.~\ref{fig:dblp_tsne} and~\ref{fig:imdb_mutag_tsne} visually demonstrate the ability of the learned representation to preserve graph structure by visualizing the the test-set latent features in a 2D t-SNE plot~\cite{van2008visualizing}. Note, in particular, how graphs that are visually similar are found nearby each other. It is evident from the t-SNE plots that the Graph2Graph model has captured semantic information of the graphs, whereas adjacency matrix embeddings (see supplementary material) fail to capture such patterns. Note also that even on a the very small trainingset of MUTAG, the embedding still preserves semantic structure. The expressiveness of the latent space embeddings is further validated on the proxy task of graph classification with limited labeled data below. More visualization results can be found in the supplementary material, including a comparison with t-SNE on the original adjacency matrices.

\subsection{Pretraining semantic representations for graph classification with limited labeled data}
\label{sec:graphClassification}

In this section, we utilize the graph autoencoder from Sec.~\ref{sec:ae} to learn semantic, low-dimensional embeddings of the dataset graphs, and apply an MLP on the latent variables for classification. In particular, we investigate the ``limited labels'' setting, which appears frequently in real life settings where data is abundant, but labeled data is scarce and expensive.

\paragraph{Problem formulation and settings.} The graph autoencoder is trained on a large training set (not using graph labels). An MLP is subsequently trained for classification on labeled training set subsamples of the latent representations, to study how the resulting model's performance depends on the size of the labeled training set. We compare our own subset model with the state-of-the-art Graph Isomorphism Network~\cite{xu2018powerful} (GIN) using similar hyperparameters (2 layers of MLP, each with hidden dimension 64; batch size 32) and selecting, for each model, the best performing epoch out of 100. Both models are given purely structural data, supplying GIN with node degrees as node labels. Both models are trained on randomly sampled subsets consisting of 0.1\%, 0.25\%, 0.5\%, 1\%, 5\%, 10\% and 100\% of the labeled training data, respectively; keeping the validation and training sets fixed. Training is repeated on new random subsets 10 times. 

\paragraph{Data.} We use the DBLP\_v1, NCI1~\cite{wale2008comparison,kim2016pubchem,shervashidze2011weisfeiler}, AIDS~\cite{aidsdata,riesen2008iam}, IMDB-BINARY~\cite{yanardag2015deep} and IMDB-MULTI datasets for classification. The rather large DBLP\_v1 is divided into 60/20/20 \% splits for training/validation/testing, whereas the remaining datasets are divided into 80/10/10 \% splits to ensure that the best models are robustly trained. These splits were kept fixed across models. 

\begin{figure}[t]
\includegraphics[width=\linewidth]{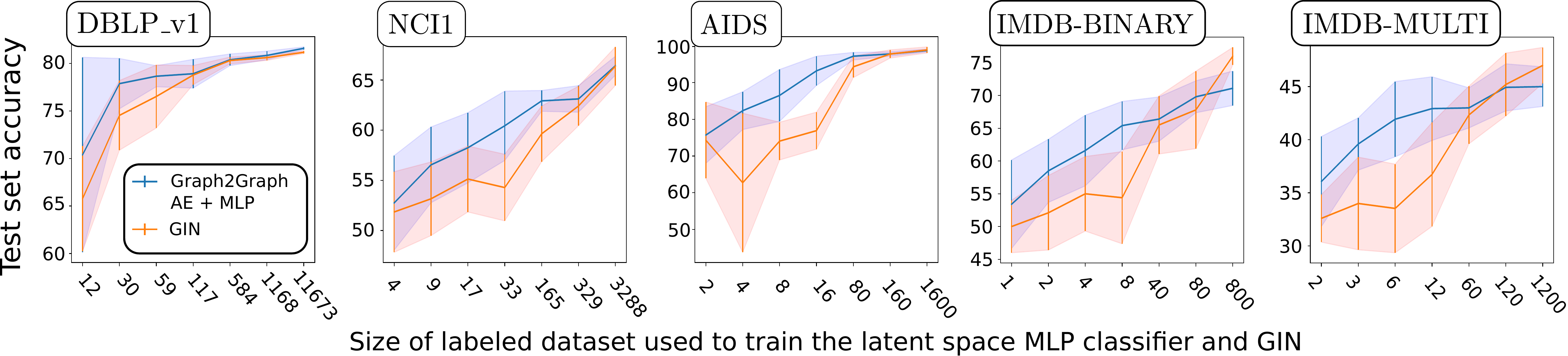}
\caption{Results from classification with features pre-trained as a Graph2Graph autoencoder.}
\label{fig:small_labels}
\end{figure}

\paragraph{Results.}
As shown in Fig.~\ref{fig:small_labels}, the model pretrained as a Graph2Graph autoencoder outperforms GIN on the small training sets, and approaches similar performance on the full training set.

%% file: discussion.tex
In this paper, we have turned our attention to graph-to-graph prediction; a problem which has so far seen very limited development within graph deep learning. We have utilized this both for graph-in, graph-out regression; a graph autoencoder; and for unsupervised pretraining of semantic representations that allow learning discriminative classification models with very little labeled data.

Our paper proposes a new, general family of problems for deep learning on graphs; namely predictions whose input and output are all graphs. We propose several tasks for learning with such models, and establish methods for validating them using publicly available benchmark data.

Our experimental validation differs from state-of-the-art in graph classification, in that we work with fixed training/validation/test splits as is commonly done in deep learning. To make this feasible, we have chosen to validate on datasets that are larger than the most commonly used graph benchmark datasets. Nevertheless, we have worked with publicly available benchmark data, making it easy for others to improve on and compare to our models. Further encouraging this, our code will be made publicly available upon publication.

While our model opens up for general solutions to new problems, it also has weaknesses. First, our current implementation assumes that all graphs have the same size, obtaining this by zero-padding all graphs to the maximal size. While this assumption is also found in other graph deep learning work, it is an expensive one, and in future work we will seek to remove it. 

Our model depends on the order of the nodes used to create the adjacency matrix, and thus per se depends on node permutation. However, in a similar fashion as~\cite{you2018graphrnn}, all graphs are represented using a depth first order before feeding them into the model, which ensures that different permutations to the input graph still gives consistent output.

The performance of the state-of-the-art benchmark GIN is lower than that reported in the literature~\cite{xu2018powerful}, for two main reasons. First, as has previously been pointed out~\cite{xu2018powerful}, the most common way to report performance for graph neural networks is by reporting the largest encountered validation performance; an approach that is explained by the widespread use of small datasets. As we have chosen to perform validation with larger datasets, we do not do this. Second, this tendency is emphasized by our use of structural information alone in order to assess differences in models rather than differences in information supplied.

This also emphasizes the potential for even stronger representations. The Graph2Graph network currently only uses structural information, and the extension to graphs with node labels or node- and edge weights, as well as graphs whose structures or attributes are stochastic, forms important directions for future work.

In conclusion, we present an autoregressive model for graph-to-graph predictions and show its utility in several different tasks, ranging from graph-valued regression, via autoencoders and their use for visualization, to graph classification with limited labeled data based on latent representations pretrained as an autoencoder.

Most work in deep learning for graphs addresses problems whose output is ``simple'', such as classification with discrete output, or regression with real-valued output. This paper demonstrates, quantitatively and visually, that graph neural networks can be used to learn far richer outputs, with corresponding rich internal representations. 